\documentclass[letterpaper]{article} 
\usepackage{aaai2027} 
\usepackage[hyphens]{url} 
\usepackage{graphicx} 
\usepackage{natbib} 
\usepackage{caption} 
\usepackage{booktabs}
\usepackage{amsfonts}

\title{FSTC-Encoder: Feature--Spatial--Temporal Correlation Learning for
Generalizable RF Sensing}
\author{
Jing Wang,
Zhu Wang,
Changlong Cheng,
Yifan Guo,
Yin Zhang
}
\affiliations{
Northwestern Polytechnical University\\
jingwang25@mail.nwpu.edu.cn, wangzhu@nwpu.edu.cn
}

\begin{document}

\maketitle

\begin{abstract}
Heterogeneous RF sensing differs substantially in feature structure,
spatial layout, and temporal scale, making existing models difficult to
reuse across devices, environments, and RF modalities. We propose
FSTC-Encoder, which unifies heterogeneous RF representation learning through
feature, spatial, and temporal correlation modeling. Structure-aware feature
encoding accommodates different signal structures, set-based spatial
encoding aggregates variable observations, and hierarchical temporal encoding
jointly captures local variations and long-range dependencies. Across sensing
tasks and modalities, FSTC-Encoder retains the same spatial--temporal backbone
architecture while varying only the feature configuration and task head.
Across Widar3.0, CSI-Bench, and XRF55, FSTC-Encoder achieves 92.15\% mean
Accuracy under multi-factor cross-domain protocols, ranks first on three of
four additional sensing tasks, remains consistently strong across WiFi,
millimeter-wave radar, and RFID, and reduces the cross-modality performance
gap from 18.85\% to 12.93\% through cross-RF learning. These results
demonstrate that FSTC-Encoder achieves high domain robustness,
task generality, and modality extensibility.
\end{abstract}

\section{Introduction}

Radio-frequency (RF) sensing transforms human-induced variations in wireless
propagation into contactless observations of the physical world. It has
expanded from human tracking, vital-sign monitoring, and gesture recognition
to daily-life understanding and increasingly diverse sensing settings
\citep{adib2014witrack,adib2015vitalradio,lien2016soli,fan2020rfdiary}.
Recent benchmarks further span multiple users, tasks, hardware platforms, and
RF modalities \citep{yang2023mmfi,huang2024wimans,wang2024xrf55,
zhu2025csibench}. This shifts the objective from optimizing one configured
pipeline to supporting changes in environments, devices, spatial layouts, and
signal modalities.

Most RF sensing models, however, target a specific sensing configuration.
Modeling-based approaches rely on selected signal quantities, whereas
learning-based approaches assume fixed input representations and encoder
structures. Both perform well under matched training and deployment conditions,
but their input construction and representation learning remain coupled to the
sensing configuration, causing substantial degradation when it changes
\citep{zhang2022widar,yan2025wisfdagr,zhang2026wicbr}. In heterogeneous RF
sensing, changes in feature composition, device configuration, or RF modality
may alter both the data distribution and the data structure.
The resulting challenge is more fundamental than conventional domain shift:
behavior-relevant correlations become entangled with the feature geometry and
spatial layout induced by a particular acquisition process.

\begin{figure}[t]
  \centering
  \includegraphics[width=0.95\columnwidth]{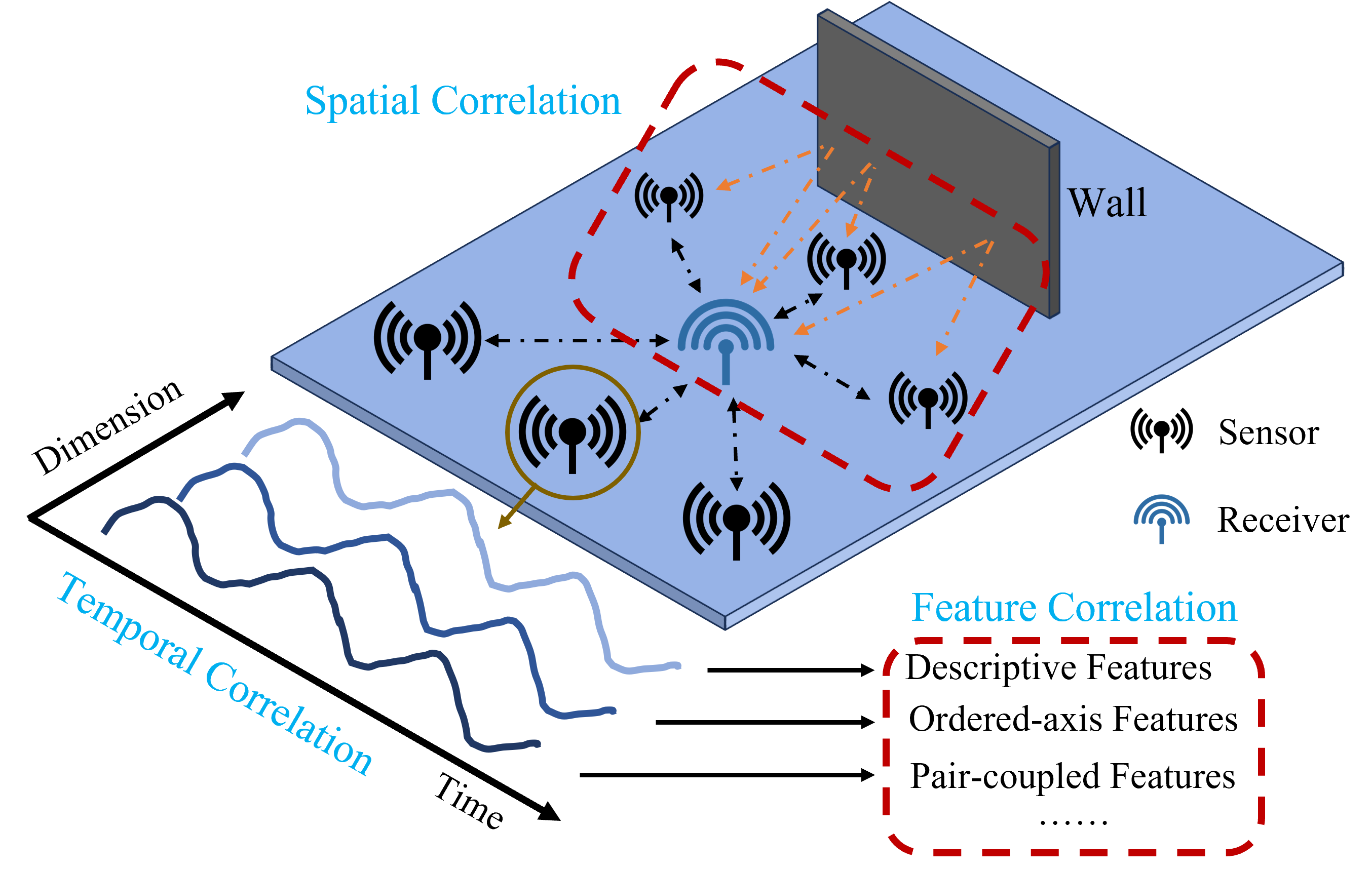}
  \caption{Feature, spatial, and temporal correlations in RF sensing. Feature
  correlation describes dependencies among signal attributes, spatial
  correlation captures complementary observations and propagation
  relationships across sensing units, and temporal correlation characterizes
  the evolution of motion-induced signal variations over time.}
  \label{fig:introduction}
\end{figure}

Despite this acquisition-level heterogeneity, human behavior induces recurring
correlations in RF signals. As illustrated in Fig.~\ref{fig:introduction},
feature correlation captures dependencies among signal attributes, spatial
correlation reflects complementary observations and propagation relationships
across sensing units, and temporal correlation describes the evolution of
motion-induced variations. These correlations provide a natural basis for
separating behavior-relevant information from configuration-specific
measurements.

Motivated by this insight, existing studies model spatial--temporal dependencies,
multidimensional relations, or correlation graphs to improve robustness
\citep{ren2023fstnet,liu2024scl,gao2026stargraph,dang2025sst}. The remaining gap
is how these correlations remain coupled to the sensing configuration. Their
operators are typically instantiated on predetermined fingerprints, tensors,
graph topologies, or modality-specific representations. Changing the feature
family, spatial cardinality, or RF modality may therefore require redefining
the input construction, relational structure, or encoder components. Flattening
heterogeneous inputs provides a common shape but discards informative feature
and spatial structures; preserving every structure with a configuration-specific
backbone prevents architectural reuse. What is missing is a factorized
architecture that preserves configuration-specific correlations where they
arise while exposing stable interfaces for reusable behavior modeling.

To address this issue, we propose \textbf{FSTC-Encoder}, a
\textbf{F}eature--\textbf{S}patial--\textbf{T}emporal
\textbf{C}orrelation Encoder that standardizes the interfaces between
correlation-encoding stages rather than standardizing the raw inputs themselves.
Structure-aware feature encoding preserves the dependency structure of each
feature family while producing a common temporal--spatial representation.
Set-based spatial encoding then removes dependence on spatial ordering and
cardinality, yielding a unified temporal sequence. With feature and spatial
variations isolated, hierarchical temporal encoding can focus on instantaneous
transitions, local motion patterns, and long-range behavioral evolution. This
factorization preserves signal-specific structure while preventing
acquisition-level changes from propagating into high-level behavior
representations. Across sensing settings, only the structure-aware feature
encoding configuration and task head change, while the spatial--temporal
backbone architecture remains unchanged.

We evaluate FSTC-Encoder on Widar3.0, CSI-Bench, and XRF55 across multiple
domains, sensing tasks, and RF modalities. It achieves 92.15\% mean Accuracy
under the multi-factor Widar3.0 protocols and obtains 52.61\% and 59.88\%
Weighted-F1 in CSI-Bench cross-environment and cross-user evaluation,
respectively. Across four additional CSI-Bench tasks, it ranks first on three
and reaches 98.65\% mean Accuracy. On XRF55, FSTC-Single achieves 85.52\% mean
Accuracy across WiFi, millimeter-wave radar, and RFID without cross-RF
training; cross-RF learning raises the mean to 87.54\% and reduces the modality
gap to 12.93\%. These results demonstrate domain robustness, task generality,
and modality extensibility in heterogeneous sensing.

Our contributions are threefold. \textbf{First,} we identify the coupling of
acquisition-specific feature geometry and spatial layout with behavior modeling
as a central obstacle to reusable RF sensing, and formulate heterogeneous
representation learning as factorized feature, spatial, and temporal
correlation encoding.
\textbf{Second,} we develop FSTC-Encoder with structure-aware feature encoding,
set-based spatial aggregation, and hierarchical temporal modeling under a
unified backbone architecture. \textbf{Third,} we systematically evaluate the
framework across domains, sensing tasks, and RF modalities, including
multi-factor cross-domain protocols.

\section{Related Work}

\subsection{Modeling-Based Wireless Sensing}

Modeling-based methods derive physically interpretable representations from
the relationship between human motion and wireless propagation. Early systems
use handcrafted RSS or CSI variations for recognition
\citep{pu2013wholehome,abdelnasser2015wigest,wang2017devicefree,
virmani2017position}, while Widar3.0 and WiHF construct more domain-stable
velocity or motion-change representations
\citep{zhang2022widar,li2022wihf}. Although effective in target
settings, these methods remain tied to selected physical quantities and
propagation assumptions, often requiring redesign across tasks, hardware, or
RF modalities.

\subsection{Learning-Based Signal Representation}

Learning-based methods replace handcrafted decision rules with neural
representation learning. Sequence models use convolutional, recurrent, or
attention-based architectures to learn local signal variations and long-term
dynamics \citep{yao2017deepsense,gu2022wigrunt,li2021crossgr}, whereas another
line reshapes CSI, radar, or RF measurements into image-like tensors and adopts
visual backbones \citep{ma2018signfi,zhao2018rfpose,li2019rfaction,
wang2019personwifi,zhao2019rfmesh,li2022unsupervised,lee2023hupr,
yan2025wisfdagr}. These generic architectures provide strong modeling capacity,
but flattening can obscure the organization of subcarriers, links, and Doppler
bins, while image conversion may introduce artificial spatial adjacency. Their
portability therefore often comes at the cost of RF-specific structure.

\subsection{Structure-Aware RF Representation Learning}

Structure-aware methods explicitly model selected correlations within RF
signals. FSTNet learns spatial--temporal dependencies from wireless
fingerprints, while Wi-CBR uses complementary feature branches
\citep{ren2023fstnet,zhang2026wicbr}. WiGNN and Star Graph encode observation
topology, whereas multidimensional CSI tensors and lightweight
spatio-spectro-temporal networks capture dependencies across signal dimensions
\citep{chen2024wignn,gao2026stargraph,du2024multidomain,dang2025sst}.

SCL is the closest work, modeling dimensional, spatial, and temporal
relationships through explicitly constructed correlation graphs across
multiple wireless modalities \citep{liu2024scl}. However, prior methods
generally bind their correlation mechanism to a fixed feature combination,
sensor topology, or graph-construction procedure. In contrast, FSTC provides
modular interfaces for heterogeneous feature structures, variable spatial
observations, and local-to-global temporal dynamics within one backbone design.

\section{Method}

\begin{figure*}[t]
    \centering
    \includegraphics[width=\textwidth]{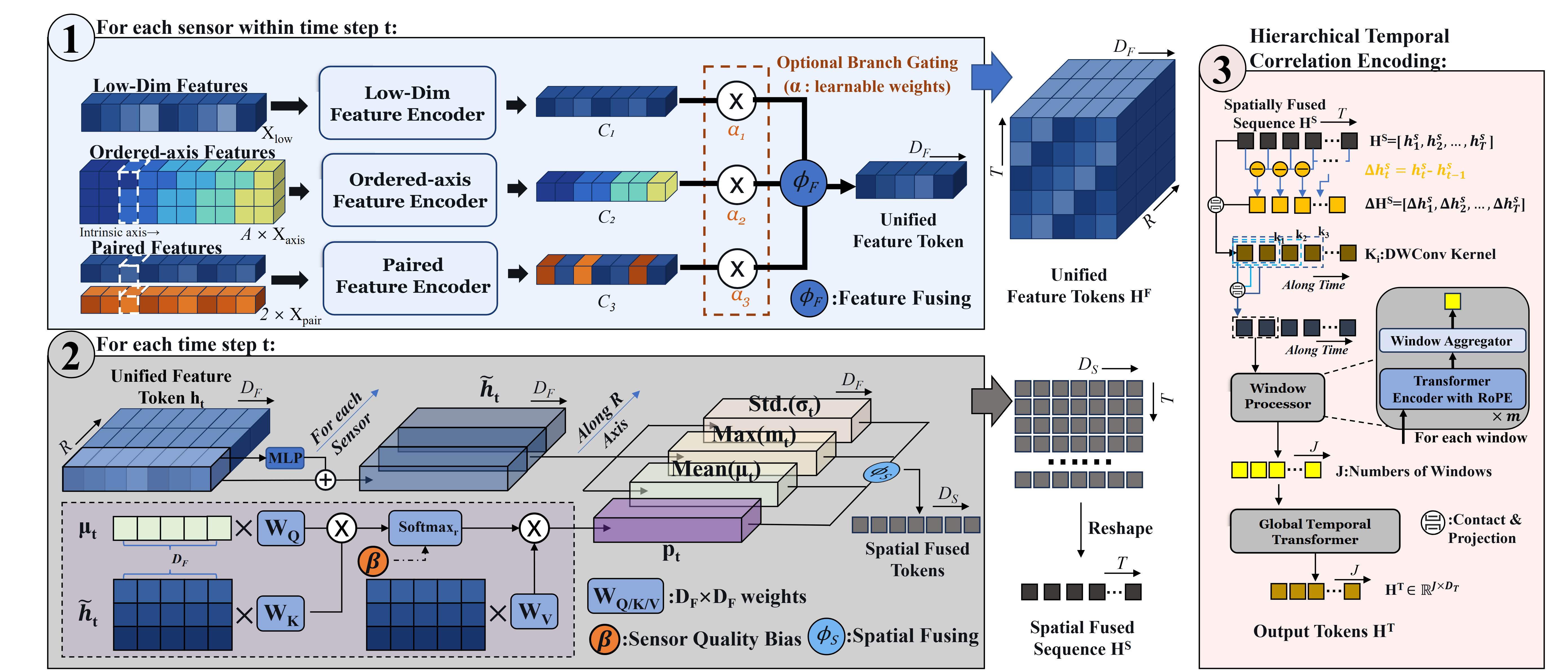}
    \caption{Overview of FSTC-Encoder. Structure-aware feature encoding maps
    heterogeneous inputs to a common interface, set-based spatial encoding
    aggregates variable observations, and hierarchical temporal encoding
    models local-to-global dynamics.}
    \label{fig:fstc_architecture}
\end{figure*}

\subsection{Problem Formulation}

We consider a sensing task $\mathcal{T}$ under an input configuration $\kappa$,
which specifies the RF modality and active feature families. User,
environment, device, location, and orientation are treated as domain factors.
For sample $i$, the heterogeneous input and task label are
\begin{equation}
\begin{array}{c}
\mathcal{X}_i^{(\kappa)}
=
\left\{
\mathbf{X}_i^{(m)} \mid m \in \mathcal{A}_\kappa
\right\},
\qquad
y_i \in \mathcal{Y}_{\mathcal{T}},
\\
\mathbf{X}_i^{(m)} \in \mathbb{R}^{T_i \times R_i \times \mathbf{D}_m}.
\end{array}
\end{equation}
where $m$ indexes feature branches rather than RF modalities and
$\mathcal{A}_\kappa$ is the active-branch set. The branch-specific internal
feature shape $\mathbf{D}_m=(D_{m,1},\ldots,D_{m,K_m})$ may contain an ordered
axis or paired component dimensions. The notation
$T_i\times R_i\times\mathbf{D}_m$ abbreviates
$T_i\times R_i\times D_{m,1}\times\cdots\times D_{m,K_m}$. Here $T_i$ is the
temporal length and $R_i=|\mathcal{R}_i|$ is the number of valid spatial
observations.
Active branches are aligned to common temporal and spatial axes within a
sample, with masks identifying invalid positions.

The predictor is defined by
\begin{equation}
\hat{\mathbf{y}}_i
=
f_{\kappa,\mathcal{T}}
\left(
\mathcal{X}_i^{(\kappa)}
\right)
\in \mathbb{R}^{C_{\mathcal{T}}},
\qquad
C_{\mathcal{T}}=|\mathcal{Y}_{\mathcal{T}}|.
\end{equation}
It must preserve branch-dependent feature structure, aggregate a variable and
unordered set of spatial observations, and model behavior over variable
temporal extents.

For domain generalization within a fixed input configuration, the source and
unseen target domains share the task label space but differ in their
class-conditional distributions:
\begin{equation}
\mathcal{Y}_s^{\mathcal{T}}
=
\mathcal{Y}_t^{\mathcal{T}},
\qquad
P_s^{(\kappa)}(\mathcal{X}\mid Y)
\neq
P_t^{(\kappa)}(\mathcal{X}\mid Y).
\end{equation}
This setting covers cross-user, cross-environment, cross-device,
cross-location, and cross-orientation evaluation. Across input configurations,
the raw feature spaces and active branches may instead differ. We therefore
require configuration-specific structure-aware feature encoding that exposes
a common interface to the shared spatial--temporal backbone
architecture. The central challenge is to prevent configuration-dependent
feature structures and observation sets from coupling to
task-level temporal behavior.

\subsection{FSTC-Encoder Overview}

As illustrated in Figure~\ref{fig:fstc_architecture}, FSTC-Encoder decomposes
RF representation learning into feature, spatial, and temporal correlation
encoding. For a fixed input configuration $\kappa$, we omit the sample index
and auxiliary masks:
\begin{equation}
\begin{array}{rl}
\mathbf{H}^{F}
&= \Phi_F^{(\kappa)}(\mathcal{X}^{(\kappa)})
\in \mathbb{R}^{T\times R\times D_F},
\\
\mathbf{H}^{S}
&= \Phi_S(\mathbf{H}^{F})
\in \mathbb{R}^{T\times D_S},
\\
\mathbf{H}^{T}
&= \Phi_T(\mathbf{H}^{S})
\in \mathbb{R}^{J\times D_T}.
\end{array}
\end{equation}
Here $T$ and $R$ denote the temporal length and the number of valid spatial
units, while $D_F$, $D_S$, and $D_T$ are the feature-fusion, spatial, and
global-token dimensions, respectively. The number of temporal windows is $J$.
The Feature Encoder maps branch-dependent internal structures to a common
$T\times R\times D_F$ interface, the Spatial Encoder removes the variable
observation axis, and the Temporal Encoder transforms the resulting sequence
into $J$ contextualized window representations. This factorization confines
feature-format variation to the feature encoding stage and variation in spatial
layout to the spatial encoding stage, allowing temporal modeling to focus on
behavior dynamics. Consequently, different input configurations change the
structure-aware feature encoding configuration while retaining the same
spatial--temporal backbone architecture.

\subsection{Structure-Aware Feature Correlation Encoding}

Stage 1 of Figure~\ref{fig:fstc_architecture} separates structure-specific
branch encoding from cross-family fusion. RF features are grouped by internal
dependency structure and processed by matching branches, which preserve the
temporal and spatial axes and output
\(
\mathbf{H}^{(m)} \in \mathbb{R}^{T \times R \times d_m}
\), standardizing internal dimensions.

\paragraph{Structure-Specific Branch Encoding.}
Low-dimensional descriptors lack meaningful local feature adjacency. We
project vectors and model short-term changes:
\begin{equation}
\begin{array}{rl}
\mathcal{P}_{\mathrm{low}}(\mathbf{X}_{\mathrm{low}})
&=
\mathrm{SiLU}\left(
\mathrm{LN}\left(
\mathbf{X}_{\mathrm{low}}\mathbf{W}_{\mathrm{low}}
+ \mathbf{b}_{\mathrm{low}}
\right)
\right),
\\
\mathbf{H}_{\mathrm{low}}
&=
\mathrm{SiLU}\left(
\mathrm{DWConv}_{t}\left(
\mathcal{P}_{\mathrm{low}}(\mathbf{X}_{\mathrm{low}})
\right)
\right)
.
\end{array}
\end{equation}
Here, $\mathrm{DWConv}_{t}$ is a depthwise one-dimensional temporal convolution
that processes projected channels independently.

Ordered-axis features, such as subcarrier or Doppler measurements, have
meaningful local adjacency along a physical axis $a$. Their branch therefore
models and aggregates axis-local patterns:
\begin{equation}
\mathbf{H}_{\mathrm{axis}}
=
\mathrm{Proj}_{m}\left(
\mathrm{Pool}_{a}^{(m)}\left(
\mathrm{AxisEnc}_{a}(\mathbf{X}_{\mathrm{axis}})
\right)
\right).
\end{equation}
Here $\mathrm{AxisEnc}_{a}$ comprises one or more one-dimensional convolutions,
normalization, and SiLU activations along $a$. Fixed-width Doppler and phase
branches use adaptive average pooling; variable or partially valid subcarrier
branches use masked mean--max pooling. The projection is identity when the
dimensions already match.

For pair-coupled features, such as real--imaginary or sine--cosine components,
the two components jointly represent a physical quantity and are embedded
before aggregation:
\begin{equation}
\begin{array}{c}
\mathbf{p}_j
=
\mathrm{SiLU}\left(
[x_j^{(1)} \Vert x_j^{(2)}]\mathbf{W}_p + \mathbf{b}_p
\right),
\\
\mathbf{H}_{\mathrm{pair}}
=
\mathrm{Agg}_{\mathrm{pair}}\left(\{\mathbf{p}_j\}\right)
\in \mathbb{R}^{T\times R\times d_{\mathrm{pair}}}.
\end{array}
\end{equation}
For fixed-size collections, $\mathrm{Agg}_{\mathrm{pair}}$ is an MLP; on a
physical axis, it adds an axis-wise convolution before aggregation.

\paragraph{Cross-Family Fusion.}
Despite different internal operators, the aligned branch outputs differ only
in feature dimension. Optional sample-level gating reweights them before they
are concatenated and jointly projected:
\begin{equation}
\mathbf{H}^{F}
=
\mathrm{SiLU}\left(
\mathrm{LN}\left(
[\mathbf{H}^{(1)} \Vert \cdots \Vert \mathbf{H}^{(M)}]
\mathbf{W}_F + \mathbf{b}_F
\right)
\right)
.
\end{equation}
Let $\mathcal{A}$ denote the set of active branches and
$d_{\Sigma}=\sum_{m\in\mathcal{A}}d_m$. Their concatenation has shape
$T\times R\times d_{\Sigma}$, and
$\mathbf{W}_F\in\mathbb{R}^{d_{\Sigma}\times D_F}$ maps it to
$\mathbf{H}^{F}\in\mathbb{R}^{T\times R\times D_F}$.
This standardizes internal dimensions without mixing temporal or spatial axes
and establishes a common interface for subsequent modeling.

\subsection{Set-Based Spatial Correlation Encoding}

Stage 2 of Figure~\ref{fig:fstc_architecture} treats receivers, links, channels,
or tags at each time step as a variable-size set. Shared observation encoding
and permutation-invariant aggregation avoid dependence on fixed ordering or
spatial layout.

\paragraph{Shared Observation Encoding.}
Each spatial unit is mapped by the same residual encoder:
\begin{equation}
\widetilde{\mathbf{h}}_{t,r}
=
\mathbf{h}^{F}_{t,r}
+
\mathrm{MLP}
\left(
\mathrm{LN}(\mathbf{h}^{F}_{t,r})
\right),
\end{equation}
Parameter sharing preserves permutation equivariance without unit-specific
encoders; $\mathcal{R}$ excludes masked observations.

\paragraph{Complementary Set Aggregation.}
We use two complementary views of the valid set. Order-invariant
statistics capture global trend, peak response, and
cross-unit variation:
\begin{equation}
\begin{array}{c}
\mu_t = \mathrm{Mean}_{r \in \mathcal{R}} \widetilde{\mathbf{h}}_{t,r},
\qquad
\mathbf{m}_t = \mathrm{Max}_{r \in \mathcal{R}} \widetilde{\mathbf{h}}_{t,r},
\\
\sigma_t = \mathrm{Std}_{r \in \mathcal{R}} \widetilde{\mathbf{h}}_{t,r}.
\end{array}
\end{equation}

In parallel, the spatial mean provides a set-level context query, while
individual observations provide keys and values. For attention head $h$, the
weight on unit $r$ is
\begin{equation}
a_{t,r}^{(h)}
=
\mathrm{Softmax}_{r\in\mathcal{R}}
\left(
\frac{
(\mathbf{W}_{q}^{(h)} \mu_t)^{\top}
(\mathbf{W}_{k}^{(h)} \widetilde{\mathbf{h}}_{t,r})
}{
\sqrt{d_h}
}
+
\beta_{r}^{(h)}
\right),
\end{equation}
where $H$ is the number of heads and $d_h=D_F/H$. The optional
$\beta_{r}^{(h)}$ is a signed additive quality bias derived from receiver-level
quality metadata and is zero when such information is unavailable. The
set-conditioned attention output is
\begin{equation}
\mathbf{p}_t
=
\mathbf{W}_{o}
\left(
\mathop{\Vert}_{h=1}^{H}
\sum_{r \in \mathcal{R}}
a_{t,r}^{(h)}
\mathbf{W}_{v}^{(h)}
\widetilde{\mathbf{h}}_{t,r}
\right).
\end{equation}

Statistics preserve the set distribution, while attention captures
sample-dependent complementarity. Their concatenation is projected into the
spatial representation:
\begin{equation}
\mathbf{h}_{t}^{S}
=
\mathrm{SiLU}
\left(
\mathrm{LN}
\left(
[\mu_t
\Vert
\mathbf{m}_t
\Vert
\sigma_t
\Vert
\mathbf{p}_t]
\mathbf{W}_S
\right)
\right).
\end{equation}
Since $\mu_t$, $\mathbf{m}_t$, $\sigma_t$, and $\mathbf{p}_t$ are in
$\mathbb{R}^{D_F}$, their concatenation is $4D_F$-dimensional. The projection
$\mathbf{W}_S\in\mathbb{R}^{4D_F\times D_S}$ yields
$\mathbf{h}_t^S\in\mathbb{R}^{D_S}$ and
$\mathbf{H}^{S}\in\mathbb{R}^{T\times D_S}$, eliminating the variable spatial
axis while preserving time.

\subsection{Hierarchical Temporal Correlation Encoding}

Stage 3 of Figure~\ref{fig:fstc_architecture} factorizes temporal correlation by
scale: full-sequence encoding may dilute short transitions, while purely local
modeling misses long-range evolution. We therefore separate motion-transition
enhancement, within-window correlation, and cross-window global modeling.

\paragraph{Motion-Transition Enhancement.}
Given $\mathbf{H}^{S}=[\mathbf{h}_{1}^{S},\ldots,\mathbf{h}_{T}^{S}]$, we
expose instantaneous changes:
\begin{equation}
\Delta \mathbf{h}_t = \mathbf{h}_{t}^{S} - \mathbf{h}_{t-1}^{S},
\qquad
\mathbf{u}_t = \mathbf{W}_{\Delta}
[\mathbf{h}_{t}^{S} \Vert \Delta \mathbf{h}_t].
\end{equation}
With $\Delta\mathbf{h}_1=0$, the projection
$\mathbf{W}_{\Delta}\in\mathbb{R}^{D_S\times2D_S}$ maps the concatenated state
and difference back to $D_S$, yielding
$\mathbf{U}=[\mathbf{u}_1,\ldots,\mathbf{u}_T]\in\mathbb{R}^{T\times D_S}$.
Parallel depthwise convolutions then capture channel-wise motion responses at
multiple durations, and a pointwise convolution fuses them:
\begin{equation}
\begin{array}{c}
\mathbf{C}^{L}
=
\mathrm{PWConv}
\left(
\mathop{\Vert}_{k \in \mathcal{K}}
\mathrm{DWConv}_{k}(\mathbf{U})
\right),
\\
\mathbf{H}^{L}
=
\mathrm{LN}
\left(
\mathbf{U}
+
\mathrm{SiLU}(\mathbf{C}^{L})
\right).
\end{array}
\end{equation}
$\mathrm{DWConv}_{k}$ is a channel-wise temporal convolution with kernel size
$k$, while $\mathrm{PWConv}$ is a $1\times1$ convolution that fuses the
multi-kernel responses across channels.

\paragraph{Within-Window Correlation.}
We project $\mathbf{H}^{L}\in\mathbb{R}^{T\times D_S}$ to
$\bar{\mathbf{H}}^{L}\in\mathbb{R}^{T\times D_L}$ and partition it into
overlapping windows of length $L_w$ and stride $S_w$. For window $j$, a shared
Local Rotary Transformer models temporal dependencies:
\begin{equation}
\begin{array}{c}
\mathbf{V}_j
=
\mathrm{LocalRoPEEnc}\left(
\bar{\mathbf{H}}_j^L,
\{0,\ldots,L_w-1\},
\mathbf{M}_j
\right),
\\
\mathbf{V}_j
=
[\mathbf{v}_{j,1},\ldots,\mathbf{v}_{j,L_w}]
\in\mathbb{R}^{L_w\times D_L}.
\end{array}
\end{equation}
Parameters are shared across windows, $\mathbf{M}_j$ masks invalid positions,
and window-relative RoPE preserves local order, making each $\mathbf{v}_{j,t}$
a context-aware motion state. A learned score measures its importance:
\begin{equation}
\begin{array}{c}
e_{j,t}=\mathbf{w}_a^{\top}\mathbf{v}_{j,t}+b_a,
\\
\alpha_{j,t}
=
\mathrm{MaskedSoftmax}_{t}\left(e_{j,t};M_{j,t}\right).
\end{array}
\end{equation}
The normalized weights form an attentive summary, paired with a masked mean:
\begin{equation}
\bar{\mathbf{v}}_{j}^{\,a}
=
\sum_{t \in \mathcal{W}_j}
\alpha_{j,t} \mathbf{v}_{j,t},
\qquad
\bar{\mathbf{v}}_{j}^{\,m}
=
\frac{
\sum_{t \in \mathcal{W}_j}
M_{j,t} \mathbf{v}_{j,t}
}{
\sum_{t \in \mathcal{W}_j}
M_{j,t}
},
\end{equation}
\begin{equation}
\mathbf{w}_j
=
\mathrm{Proj}
\left(
\bar{\mathbf{v}}_{j}^{\,a}
\Vert
\bar{\mathbf{v}}_{j}^{\,m}
\right),
\end{equation}
Attention emphasizes salient motion states, whereas the mean preserves the
complete local context. Their $2D_L$-dimensional concatenation is projected to
$\mathbf{w}_j\in\mathbb{R}^{D_T}$.

\paragraph{Cross-Window Physical-Time Modeling.}
To model long-range evolution, let $\tau_j$ be the physical timestamp at the
center of window $j$ and $\delta$ a fixed temporal normalization unit. The
global positional coordinate is $p_j=\tau_j/\delta$.
The Global Rotary Transformer then operates on the window sequence:
\begin{equation}
\mathbf{H}^{T}
=
\mathrm{GlobalRoPEEnc}
\left(
\{\mathbf{w}_j\}_{j=1}^{J},
\{p_j\}_{j=1}^{J},
\mathbf{M}^{W}
\right).
\end{equation}
Here $p_j\in\mathbb{R}$ is a scalar RoPE coordinate and
$\mathbf{M}^{W}\in\{0,1\}^{J}$ is the valid-window mask. Both the input token
sequence and $\mathbf{H}^{T}$ are in $\mathbb{R}^{J\times D_T}$. Unlike an
additive embedding, $p_j$ directly preserves the physical spacing between
windows. When timestamps are unavailable, $\tau_j$ is recovered from frame
indices and the nominal sampling interval. The temporal path maps
$T\times D_S$ frame states to $J\times D_T$ window tokens, retaining
local motion detail and modeling long-range evolution.

\paragraph{Information-Rich Output Tokens.}
The resulting sequence
$\mathbf{H}^{T}=[\mathbf{h}_{1}^{T},\ldots,\mathbf{h}_{J}^{T}]
\in\mathbb{R}^{J\times D_T}$ is the output of FSTC-Encoder. Because these
tokens are produced through progressive feature, spatial, and temporal
correlation modeling, they jointly encode structure-specific signal cues,
complementary observation evidence, contextualized local motion, and
cross-window behavioral evolution. This information-rich token interface can
support different sensing tasks through lightweight aggregation, projection,
and task heads without changing the FSTC backbone.

For these tasks, attention-weighted statistical pooling gives
\begin{equation}
\mathbf{z}
=
\mu_a
\Vert
\sigma_a
\Vert
\mathbf{h}_{\max},
\end{equation}
where $\mu_a$ and $\sigma_a$ are the attention-weighted mean and standard
deviation over valid tokens, and $\mathbf{h}_{\max}$ is their element-wise
maximum. Thus $\mathbf{z}\in\mathbb{R}^{3D_T}$. The projection maps it to
$\mathbb{R}^{D_Z}$, and
$g_{\mathcal{T}}:\mathbb{R}^{D_Z}\rightarrow\mathbb{R}^{C_{\mathcal{T}}}$
produces the task logits:
\begin{equation}
\hat{\mathbf{y}}
=
g_{\mathcal{T}}
\left(
\mathrm{Proj}(\mathbf{z})
\right).
\end{equation}

\begin{table*}[t]
\centering
\normalsize
\setlength{\tabcolsep}{3pt}
\renewcommand{\arraystretch}{1.05}
\begin{tabular*}{\textwidth}{@{\extracolsep{\fill}}p{0.38\textwidth}p{0.21\textwidth}ccccc@{}}
\toprule
Method & Processing Flow & \multicolumn{3}{c}{Widar3.0} & Mean & XRF55 \\
\cmidrule(lr){3-5} \cmidrule(lr){7-7}
 & & CL & CO & CE & & CS \\
\midrule
Widar3.0~\cite{zhang2022widar} & CSI $\rightarrow$ DFS $\rightarrow$ BVP & 90.48 & 81.58 & 83.30 & 85.12 & -- \\
WiHF~\cite{li2022wihf} & CSI $\rightarrow$ DFS $\rightarrow$ MCP & 91.22 & 80.64 & -- & -- & -- \\
WiSR~\cite{liu2024wisr} & CSI image & 67.73 & 69.74 & 52.77 & 63.41 & 26.66 \\
Recurrent ConFormer~\cite{shang2023conformer} & Raw CSI & 73.84 & 85.88 & 50.38 & 70.03 & 16.54 \\
Wi-SFDAGR~\cite{yan2025wisfdagr} & CSI $\rightarrow$ Phase & 97.30 & \textbf{97.17} & 95.52 & 96.66 & 57.99 \\
\textbf{Wi-CBR (easy)}~\cite{zhang2026wicbr} & CSI $\rightarrow$ Phase, DFS & \textbf{98.34} & 96.30 & \textbf{96.87} & \textbf{97.17} & \textbf{66.05} \\
\midrule
Wi-CBR (hard)~\cite{zhang2026wicbr} & CSI $\rightarrow$ Phase, DFS & 92.98 & 89.01 & 85.60 & 89.20 & 40.31 \\
PatchTST~\cite{nie2023patchtst} & CSI $\rightarrow$ DFS & 87.60 & 72.74 & 74.38 & 78.24 & 22.29 \\
\textbf{FSTC-Encoder (ours)} & CSI $\rightarrow$ RSSI, DFS, dCSI & \textbf{93.84} & \textbf{92.43} & \textbf{90.17} & \textbf{92.15} & \textbf{41.11} \\
\bottomrule
\end{tabular*}
\caption{Results on Widar3.0 and XRF55. All values are Accuracy
(\%). Rows above and below the horizontal rule use the official easy and our
harder multi-factor protocols, respectively. Bold values mark the
best result within each protocol group.}
\label{tab:widar3_results}
\end{table*}

\begin{table*}[t]
\centering
\small
\setlength{\tabcolsep}{4pt}
\renewcommand{\arraystretch}{1.08}
\begin{tabular*}{\textwidth}{@{\extracolsep{\fill}}lcccccc@{}}
\toprule
Method & ID & CDev & CE & CU & CD Avg. & Worst CD \\
\midrule
MLP~\cite{rumelhart1986backprop} & 83.31 & 50.79 & 43.45 & 42.05 & 45.43 & 42.05 \\
LSTM~\cite{hochreiter1997lstm} & 94.54 & 57.04 & 46.22 & 45.70 & 49.65 & 45.70 \\
ResNet-18~\cite{he2016resnet} & 94.13 & \underline{63.57} & 50.90 & 52.07 & 55.51 & 50.90 \\
Transformer~\cite{vaswani2017attention} & 94.51 & 57.80 & 47.17 & 46.67 & 50.55 & 46.67 \\
ViT~\cite{dosovitskiy2021vit} & \underline{95.67} & \textbf{63.65} & \underline{51.86} & 51.48 & \underline{55.66} & \underline{51.48} \\
PatchTST~\cite{nie2023patchtst} & 94.67 & 58.05 & 49.55 & 49.25 & 52.28 & 49.25 \\
TimeSformer-1D~\cite{bertasius2021timesformer} & 95.05 & 55.70 & 46.63 & 45.74 & 49.36 & 45.74 \\
Wi-CBR~\cite{zhang2026wicbr} & \textbf{96.80} & 45.95 & 44.52 & \underline{58.06} & 49.51 & 44.52 \\
\textbf{FSTC-Encoder (ours)} & 93.32 & 61.32 & \textbf{52.61} & \textbf{59.88} & \textbf{57.94} & \textbf{52.61} \\
\bottomrule
\end{tabular*}
\caption{Human activity recognition results on CSI-Bench. All metrics are
Weighted-F1 (\%). Bold and underlined values indicate the best and second-best
results, respectively. ID, CDev, CE, and CU denote in-domain, cross-device,
cross-environment, and cross-user, respectively. CD denotes cross-domain; CD
Avg. and Worst CD are the mean and minimum of its three protocols.}
\label{tab:csibench_results}
\end{table*}

\section{Experiments}

We test four hypotheses across Widar3.0, CSI-Bench, and XRF55. \textbf{First,}
Feature--Spatial--Temporal correlation modeling remains robust under shifts in
location, orientation, environment, user, and device. \textbf{Second,} the same
backbone architecture supports sensing tasks with different dynamics and
prediction objectives. \textbf{Third,} modality-specific structure-aware
feature encoding supports different RF modalities without redesigning
the spatial--temporal backbone. \textbf{Finally,} the performance gains cannot
be attributed to model scale or richer inputs alone.

\subsection{Experimental Setup}

\paragraph{Widar3.0.}
Widar3.0~\cite{zhang2022widar} is a WiFi gesture-recognition benchmark whose
official splits are close to saturation, with recent methods approaching
$100\%$ Accuracy~\cite{zhang2026wicbr}. We therefore use the complete dataset
and stricter multi-fold cross-location (CL), cross-orientation (CO), and
cross-environment (CE) protocols, holding out each condition of the target
factor in turn while allowing the remaining factors to vary.

\paragraph{CSI-Bench.}
CSI-Bench~\cite{zhu2025csibench} is a real-world multi-task WiFi benchmark
collected with commodity electronic devices rather than dedicated sensing
boards. For HAR, we follow the official in-domain, cross-device,
cross-environment, and cross-user splits to evaluate domain generalization.
We use official in-domain splits for Fall Detection,
Breathing Detection, Room-Level Localization, and Motion Source Recognition
to evaluate task generality.

\paragraph{XRF55.}
XRF55~\cite{wang2024xrf55} provides synchronized WiFi, millimeter-wave radar,
and RFID measurements. We follow its recommended four-fold, 8-gesture WiFi
cross-scene (CS) protocol and the official 55-class HAR protocol across all three
RF modalities.

\paragraph{Baselines and Metrics.}
We compare three main baseline families: RF sensing-specific methods such as
Wi-CBR; general sequence models including MLP, LSTM, Transformer, PatchTST, and
TimeSformer-1D; and general image models including ResNet-18 and ViT. Following
the official evaluation settings, we report Accuracy for Widar3.0 and XRF55
and primarily Weighted-F1 for CSI-Bench.

\subsection{Cross-Domain Generalization}

\paragraph{Cross-Domain Generalization on Widar3.0.}
The upper block of Table~\ref{tab:widar3_results} first confirms why a harder
protocol is needed: Wi-CBR nearly saturates the official splits. When the
nuisance factors vary under our multi-fold evaluation, FSTC-Encoder achieves
the highest mean among the hard-protocol methods and remains above $90\%$ on
CL, CO, and CE. Its $2.95\%$ mean improvement over Wi-CBR provides the central
comparison, while the advantage over PatchTST shows that the result holds
against both RF-specific and general sequence designs. The key outcome is
therefore stable adaptation across multiple domain factors rather than a peak
score on a saturated split.

\paragraph{In-the-Wild Domain Shifts on CSI-Bench.}

Table~\ref{tab:csibench_results} tests naturally occurring device,
environment, and user shifts rather than controlled single-factor changes.
FSTC-Encoder ranks first in CD Avg. and Worst CD and also leads on CE and CU.
This joint result is more informative than any single split: several baselines
with stronger ID or CDev performance become less reliable under unseen
environments and users, whereas FSTC-Encoder maintains the strongest lower
bound and the most balanced cross-domain profile. Its weaker CDev result may
partly reflect information loss when reconstructing the multi-link structure
from the flattened official release. Overall, the results demonstrate that the
learned representation remains robust across different naturally occurring
shifts rather than specializing in one domain factor.

\paragraph{Cross-Scene Evaluation on XRF55.}
XRF55 provides a single cross-domain setting based on cross-scene evaluation.
Under its official four-fold split, FSTC-Encoder achieves the strongest result
among methods evaluated under the same protocol, slightly exceeding the
RF-specific Wi-CBR and clearly outperforming the general PatchTST. Together
with the Widar3.0 and CSI-Bench results, this shows that the generalization
benefit persists across both controlled multi-factor shifts and naturally
occurring cross-scene variation.

\subsection{Task Generality across Sensing Tasks}

\begin{table}[t]
\centering
\small
\setlength{\tabcolsep}{3pt}
\renewcommand{\arraystretch}{1.06}
\begin{tabular}{@{}lcccc@{}}
\toprule
Method & Fall & Breath. & Local. & MSR \\
\midrule
MLP & 92.16 & 97.59 & 87.14 & 98.86 \\
ResNet-18 & 94.88 & 98.58 & \textbf{100.00} & \underline{99.56} \\
LSTM & \underline{94.93} & 98.62 & 99.12 & 98.42 \\
Transformer & 94.28 & 98.64 & 99.27 & 98.61 \\
ViT & 93.58 & 98.63 & \underline{99.94} & 98.74 \\
PatchTST & 94.03 & \underline{98.84} & 99.91 & 98.86 \\
TimeSformer-1D & 93.86 & 98.68 & \textbf{100.00} & 98.38 \\
\textbf{FSTC-Encoder (ours)} & \textbf{95.82} & \textbf{99.31} & 99.81 & \textbf{99.66} \\
\bottomrule
\end{tabular}
\caption{Accuracy (\%) on four CSI-Bench tasks: Fall Detection (Fall),
Breathing Detection (Breath.), Room-Level Localization (Local.), and Motion
Source Recognition (MSR). Bold and underlined values mark the best and
second-best.}
\label{tab:csibench_multitask_results}
\end{table}

\begin{table*}[t]
\centering
\normalsize
\setlength{\tabcolsep}{6pt}
\renewcommand{\arraystretch}{1.08}
\begin{tabular}{llcccc}
\toprule
Method & Training Paradigm & WiFi & mmWave & RFID & Mod. Gap $\downarrow$ \\
\midrule
DML~\cite{wang2024xrf55} & Cross-RF mutual learning & 87.26 & 86.88 & 56.82 & 30.44 \\
WiTalk~\cite{yang2025witalk} & DML + text semantics & 91.16 & 87.34 & 59.41 & 31.75 \\
X-Fi~\cite{chen2025xfi} & Unified multimodal training & 55.70 & 83.90 & 42.50 & 41.40 \\
PTA~\cite{weng2026pta} & Multimodal teacher distillation & 82.34 & 90.03 & 55.04 & 34.99 \\
COMPASS~\cite{wang2026compass} & Proxy-token cross-modal learning & 85.40 & 90.50 & 48.40 & 42.10 \\
\textbf{FSTC-Single (ours)} & Independent unimodal training &
\underline{92.06} & \textbf{91.28} & \underline{73.21} & \underline{18.85} \\
\textbf{FSTC-Multi (ours)} & Cross-RF DML training &
\textbf{92.21} & \underline{91.12} & \textbf{79.28} & \textbf{12.93} \\
\bottomrule
\end{tabular}
\caption{Single-modality inference results on XRF55 under different training
paradigms. All metrics are Accuracy (\%). Bold and underlined values indicate
the best and second-best results, respectively. Mod. Gap is the difference
between the best and worst modality results for each method; lower is better.}
\label{tab:xrf55_multirf_results}
\end{table*}

\begin{table}[t]
\centering
\small
\setlength{\tabcolsep}{1.5pt}
\renewcommand{\arraystretch}{1.06}
\begin{tabular}{@{}p{0.18\columnwidth}p{0.25\columnwidth}cc@{}}
\toprule
Method & Feature-family & Weighted-F1 & $\Delta$ Weighted-F1 \\
\midrule
FSTC & $A+D+\Delta$ & \textbf{93.47} & -- \\
 & $D+\Delta$ & 79.01 & $-14.46$ \\
 & $D$ & 63.26 & $-30.21$ \\
 & $\Delta$ & 72.42 & $-21.05$ \\
\midrule
PatchTST & $A+D+\Delta$ & 50.23 & -- \\
 & $D$ & \textbf{85.74} & $+35.51$ \\
 & $\Delta$ & 13.83 & $-36.40$ \\
\bottomrule
\end{tabular}
\caption{Feature-family ablation on CSI-Bench HAR. $A$, $D$, and $\Delta$
denote the low-dimensional amplitude, dynamic-residual, and temporal-difference
families.}
\label{tab:input_config_ablation}
\end{table}

\begin{table}[t]
\centering
\small
\setlength{\tabcolsep}{2.5pt}
\renewcommand{\arraystretch}{1.06}
\begin{tabular}{@{}p{0.40\columnwidth}cccc@{}}
\toprule
Variant & CL & CO & CE & Avg. Drop \\
\midrule
FSTC-Encoder & \textbf{93.84} & \textbf{92.43} & \textbf{90.17} & -- \\
w/o Feature Encoder & 90.32 & 88.72 & 86.37 & 3.68 \\
w/o Spatial Encoder & 87.75 & 84.46 & 81.88 & 7.45 \\
w/o Temporal Encoder & 71.29 & 69.47 & 65.48 & 23.40 \\
\bottomrule
\end{tabular}
\caption{Structural ablation on Widar3.0. All values are Accuracy (\%). Avg.
Drop is the mean decrease across CL, CO, and CE relative to full FSTC.}
\label{tab:structural_ablation}
\end{table}

The four CSI-Bench tasks require different forms of evidence, ranging from
abrupt events and weak periodic motion to spatial-state discrimination and
motion-source attribution. Despite these different dynamics and prediction
objectives, the FSTC backbone architecture remains unchanged. FSTC-Encoder
ranks first on three tasks and remains within $0.19\%$ of the best result on
the already saturated localization task. More importantly, its $98.65\%$ mean
Accuracy is the highest across all four tasks. This consistent performance,
rather than an isolated gain on one task, shows that the same
Feature--Spatial--Temporal organization provides a general representation for
heterogeneous sensing objectives.

\subsection{Modality Extensibility across RF Modalities}

Table~\ref{tab:xrf55_multirf_results} compares single-modality inference under
different training paradigms on XRF55. FSTC retains the same backbone
architecture across WiFi, mmWave, and RFID and changes only the structure-aware
feature encoding configuration. Even without cross-RF training, FSTC-Single
ranks among the top two methods on every modality, reaches $85.52\%$ mean
Accuracy, and limits the modality gap to $18.85\%$. This consistency shows that
the architecture adapts across RF sensing mechanisms without favoring one
modality.

Cross-RF DML further raises the mean Accuracy to $87.54\%$ and narrows the
modality gap to $12.93\%$, primarily by improving the weaker RFID modality while
preserving WiFi and mmWave performance. Cross-RF learning therefore transfers
useful knowledge without degrading the stronger modalities.

\subsection{Ablation Studies}

We distinguish gains from richer inputs and model capacity, then isolate each
encoding stage's contribution.

\paragraph{Input Configuration Ablation.}
Table~\ref{tab:input_config_ablation} compares FSTC-Encoder and PatchTST under
matched feature-family definitions; $\Delta$ Weighted-F1 is measured from each
method's full $A+D+\Delta$ configuration.

FSTC-Encoder reaches $93.47\%$ with all three families, improving by $14.46\%$
over $D+\Delta$. PatchTST instead performs best with $D$ alone and falls by
$35.51\%$ when all families are combined. Thus, the gain depends on
structure-aware encoding and correlation learning, not input quantity alone.

Model scale also does not explain this contrast. FSTC-Encoder contains 2.019M
parameters, remains close in size to PatchTST and ViT, and uses $40\%$--$91\%$
fewer parameters than the larger Transformer, ResNet-18, and Wi-CBR models. The
input and scale controls therefore attribute the gain to structured
feature--spatial--temporal modeling rather than greater capacity.

\paragraph{Structural Ablation.}
Removing feature, spatial, or temporal encoding reduces the mean Accuracy by
$3.68\%$, $7.45\%$, and $23.40\%$, respectively, across the Widar3.0 protocols.
All stages are complementary, with temporal encoding contributing most. Each
removal also causes its largest loss under CE, showing that the full
factorization becomes more important under stronger environmental shifts.

\section{Conclusion}

FSTC-Encoder factorizes heterogeneous RF sensing into feature, spatial, and
temporal correlation encoding. Across three datasets, it demonstrates robust
domain, task, and modality adaptability, including cross-RF learning gains.
Matched-input, scale, and structural analyses attribute these gains to the
architecture rather than richer inputs or larger models.

\bibliography{references}

\end{document}